\documentclass[letterpaper, 10 pt, conference]{ieeeconf}  

\IEEEoverridecommandlockouts                              
\usepackage[usenames, dvipsnames]{color}
\usepackage[noadjust]{cite}
\usepackage{booktabs}
\usepackage{graphicx}
\usepackage{caption}
\usepackage{subcaption}
\usepackage{float}
\usepackage{amsmath}
\usepackage{algorithmicx}
\usepackage{subfiles}
\usepackage{multirow}
\usepackage{mathrsfs}
\graphicspath{{images}{../images/}}
\usepackage{graphicx}
\usepackage[utf8]{inputenc}
\usepackage{amsmath}
\usepackage[hidelinks]{hyperref}

\usepackage[usenames,dvipsnames,table]{xcolor}
\usepackage{soul,xcolor}
\usepackage{siunitx}
\usepackage[export]{adjustbox}
\usepackage{multicol}
\usepackage{tikz}
\definecolor{taga}{RGB}{0,0,255}
\definecolor{tagb}{RGB}{0,67,169}
\definecolor{tagc}{RGB}{255,101,0}
\definecolor{tagd}{RGB}{255,0,0}
\definecolor{route}{RGB}{0,248,0}

\title{\LARGE \bf
Allo-centric Occupancy Grid Prediction for Urban Traffic Scene Using Video Prediction Networks\\
}

\author{Rabbia Asghar$^{1}$, Lukas Rummelhard$^{1}$, Anne Spalanzani$^{1}$, Christian Laugier$^{1}$\\
\thanks{$^{1}$~Univ. Grenoble Alpes, Inria, 38000 Grenoble, France, email:  FirstName.LastName@inria.fr}%
}

\begin{document}
\setstcolor{red}

\maketitle
\thispagestyle{empty}
\pagestyle{empty}
\setlength{\belowdisplayskip}{2pt}
\widowpenalty10000
\clubpenalty10000
\addtolength{\abovedisplayskip}{-5pt}

\begin{abstract}
Prediction of dynamic environment is crucial to safe navigation of an autonomous vehicle. Urban traffic scenes are particularly challenging to forecast due to complex interactions between various dynamic agents, such as vehicles and vulnerable road users.
Previous approaches have used ego-centric occupancy grid maps to represent and predict dynamic environments. However, these predictions suffer from blurriness, loss of scene structure at turns, and vanishing of agents over longer prediction horizon. In this work, we propose a novel framework to make long-term predictions by representing the traffic scene in a fixed frame, referred as allo-centric occupancy grid. This allows for the static scene to remain fixed and to represent motion of the ego-vehicle on the grid like other agents'. We study the allo-centric grid prediction with different video prediction networks and validate the approach on the real-world Nuscenes dataset. The results demonstrate that the allo-centric grid representation significantly improves scene prediction, in comparison to the conventional ego-centric grid approach. 


\end{abstract}


\section{INTRODUCTION} \label{sec:introduction}


Prediction of traffic scene evolution is essential to an autonomous vehicle for planning as well as detecting dangerous situations. 
In urban traffic scenarios, the vehicles not only interact with other vehicles, but also share space with vulnerable road users such as pedestrians and cyclists. Key challenges involve the uncertainty and multi-modality of the behaviour of agents in the environment, and complex multi-agents interactions \cite{mozaffari2020deep}. 
While human drivers show superior ability to forecast the agents' behaviour and interactions in such traffic scenes, it remains a challenge for autonomous vehicles. 

\begin{figure*}[h]
	\centering
	    \includegraphics[trim={0 5 0 0},clip,width=1.8\columnwidth]{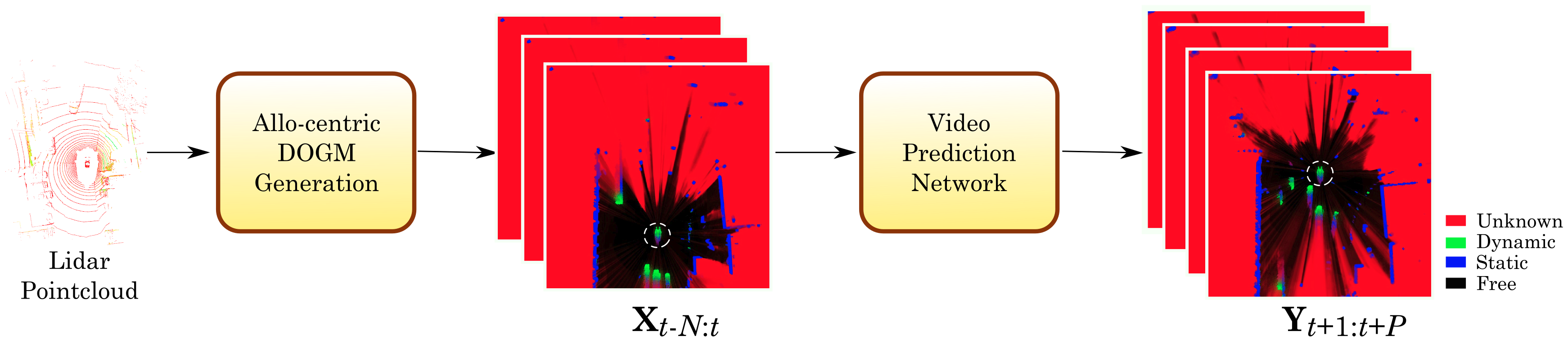}
\caption{Overview of our proposed approach. The allo-centric DOGM is represented as an image. Each channel red, green and blue represent unknown, dynamic and static cells respectively. The black space represents known free space. The ego-vehicle is circled in dotted line in both input and target output sequences.}
\label{fig:overview}
\end{figure*}

Data-driven methods provide powerful tools to solve prediction problems, particularly dealing with complex social interactions \cite{alahi2016social}. Most conventional approaches are object or agent-based and rely on heavily pre-processed data \cite{nuscenes2019}, \cite{ettinger2021large}. Dynamic Occupancy Grip Maps (DOGMs), on the other hand, allow for end-to-end learning due to their discretized spatial representation, without higher-level segmentation \cite{negre2014hybrid}. Additionally, DOGMs are versatile in terms of sensor dependency, and can be generated from a variety of raw sensor data, such as Lidar or camera images.
In our work, we use Bayesian-filter-based DOGM \cite{rummelhard2015conditional} that provide us with a spatially-dense model representation of static and dynamic space, as well as free and unknown space in the environment, as shown in Fig\ref{fig:overview}. 

As the DOGM is generated using data from the vehicle-mounted sensors,
the grid is traditionally ego-centric,i.e. the position of ego-vehicle is fixed in the grid. 
While this is an effective method in scene representation, it complicates the long-term prediction problem. For a dynamic ego-vehicle, the complete scene translates and/or rotates around the ego-vehicle, even the static components in the scene. Therefore, the prediction network must transform every cell in the grid, leading to blurry and vanishing static scene at longer prediction time horizons.

To address this, we instead generate DOGMs with respect to a fixed reference frame, referred as \textbf{allo-centric grid}. While the observed space around the ego-vehicle remains the same, the static scene structure in the allo-centric grid remains fixed.
This is illustrated in Fig. \ref{fig:overview} where the ego-vehicle is encircled, the vehicle moves like other agents in the scene. 

We approach the long-term multi-step predictions of allo-centric DOGM as a video prediction problem due to the inherent similarities between an image and an occupancy grid, and both being a spatio-temporal problem  \cite{oprea2020review}. 
Results incorporating different video prediction networks are studied, including state-of-the-art recurrent neural networks and memory-augmented network approaches.
We compare and evaluate the prediction results of allo-centric and ego-centric grids for identical scenes and demonstrate the superior performances of the allo-centric grid predictions.


The proposed approach is validated with the real-world NuScenes dataset \cite{nuscenes2019} of urban traffic scenes. We show that allo-centric grids significantly improve the prediction results and demonstrate the ability to retain the scene structure and learn behaviours.

The paper is organized as follows. Section \ref{sec:related_work} discusses related work to video and scene predictions. Section \ref{sec:approach} describes
the system overview. Section \ref{sec:implementation} and \ref{sec:results} present implementations, results and analysis. Finally conclusions are drawn in section \ref{sec:conclusion}.
\section{Related Work} \label{sec:related_work}
\subsection{Video Prediction}\label{sec:related_work_a}
Spatio-temporal deep-learning methods have been effectively used for video prediction problems. Commonly, combinations of Convolutional Neural Networks (CNNs) and Recurrent Neural Networks (RNNs) are incorporated. CNNs are capable of extracting spatial information and capturing inter-dependencies of the surrounding pixels while RNNs, such as long short-term memory (LSTM) blocks, capture the sequential or temporal dependencies. 
Lotter et al. proposed Predictive Coding Network (PredNet), a deep learning network architecture that comprises of vertically-stacked Convolutional LSTMs (ConvLSTMs) where the local error and the prediction signal are propagated bottom-up and top-down respectively \cite{Lotter2017}.
Wang et al. addresses the video prediction challenges of capturing short-term and long-term dynamics with the PredRNN architecture \cite{9749915}. Building on their original approach \cite{NIPS2017_e5f6ad6c}, they introduce memory-decoupled spatio-temporal LSTM (ST-LSTM) blocks, feature zigzag memory flow and a novel curriculum learning strategy to improve prediction results.
Kim et al. takes inspiration from memory-augmented networks to use external memory (LMC-Memory) to learn and store long-term motion dynamics and propose a memory query decomposition to address the high-dimensionality of motions in video predictions 
\cite{lee2021video}.

\subsection{Occupancy Grid Prediction}
Jeon et al. proposed conventional ConvLSTM to predict interaction-intensive traffic scenes on occupancy grids \cite{8500567}. 
The approach represents only vehicles in the occupancy grid, their states extracted from camera inputs.
Desquaire et al. \cite{Dequaire2017}, proposed an end-to-end object-tracking approach by incorporating directly Lidar sensor data to predict the binary grid, using recurrent neural network. To incorporate ego-vehicle motion, they utilize a spatial transformer to allow internal memory of RNNs to learn environment of the state.
Mohajerin et al. \cite{mohajerin2019multi} suggested an RNN-based architecture with a difference learning method, 
and makes OGM prediction in the field of view of ego-vehicle front camera. 
Schreiber et al. \cite{schreiber2019long} proposed an encoder-decoder network architecture, along with skip connections, to make long-term DOGM predictions.
While they collect the sensor data from an autonomous vehicle, the vehicle remains stationary and only acts as the sensor collection point at different intersections.
Itkina et al. proposed to use evidential occupancy grid and implement PredNet architecture for the prediction \cite{itkina2019dynamic}. The approach is then carried forward to develop the double-pronged architecture \cite{Toyungyernsub2020a} and attention-augmented ConvLSTM  \cite{lange2020attention}. The latter work is able to make long-term predictions, however at turns the predictions still lose the scene structure. Mann et al. \cite{https://doi.org/10.48550/arxiv.2205.03212} addressed the problem of OGM prediction in urban scenes by incorporating vehicles semantics in the environment. Their proposed method depends on the annotated vehicle data labels available in the dataset.

Contrary to the conventional Occupancy Grid Prediction, we present an allo-centric DOGM representation to predict the urban traffic scene with respect to a fixed reference frame.
Apart from the conventional recurrent representation learning approaches, we also use memory-augmented learning-based video-prediction method, in relevance to learning long-term motion context of the dynamic agents.


\section{System Overview} \label{sec:approach}
We discuss here the overall proposed approach for allo-centric DOGM prediction, the pipeline is summarized in Fig. \ref{fig:overview}.

\subsection{Dynamic Occupancy Grid Maps}\label{sec:approach-a}
Dynamic occupancy grid maps provide a discretized representation of environment in a bird's eye view, where every cell in the grid is independent and carries information about the associated occupancy and velocity.

\begin{figure}[h!]
\centering
\includegraphics[width=\columnwidth]{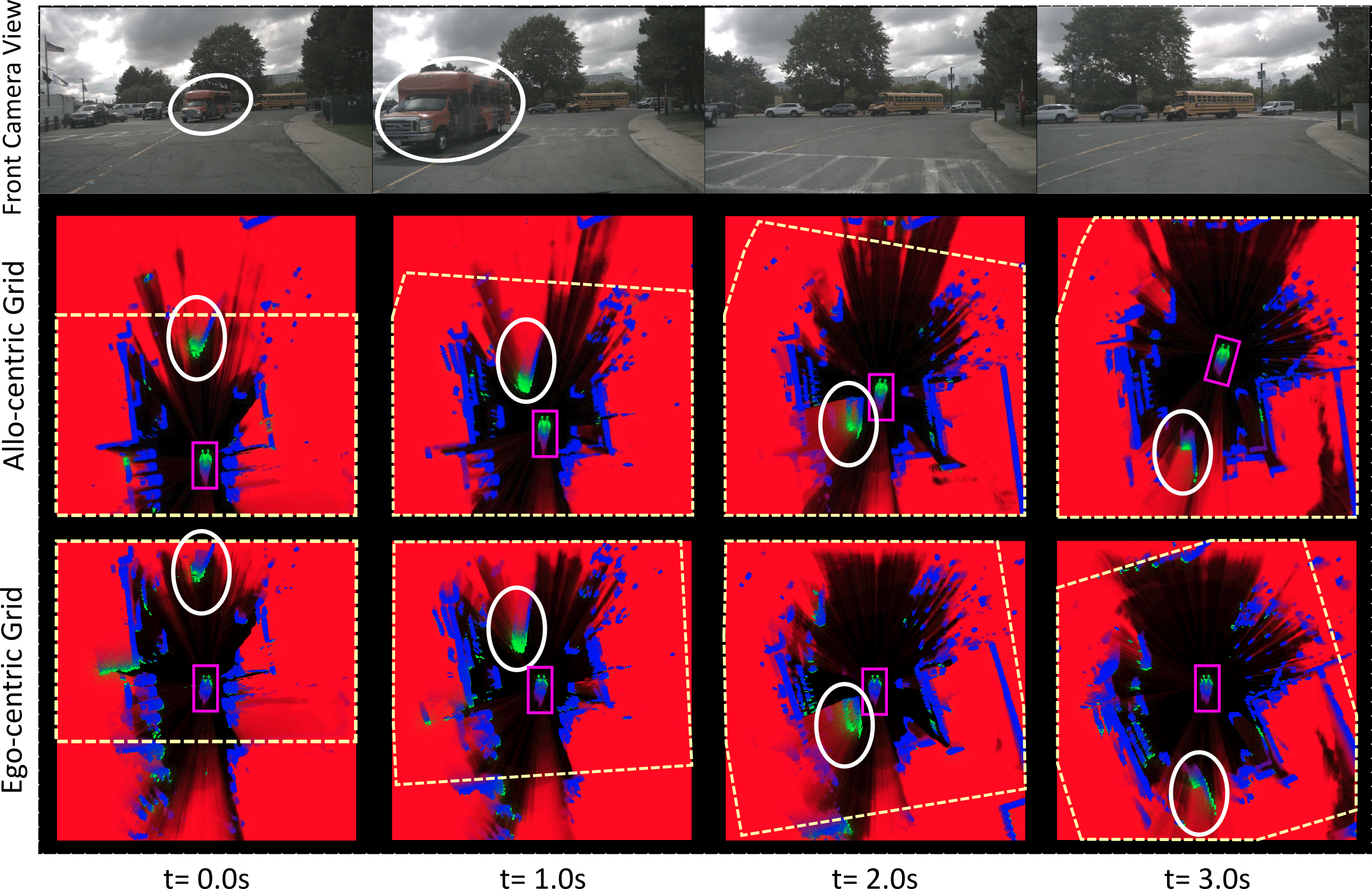}
\caption{Visualization of allo-centric and ego-centric grids, generated for the same scene. The area marked by yellow lines is the common region covered by both grids up until the $t$-th sequence. The ego-vehicle is boxed in pink grid and the bus passing by is encircled in white.}
\label{fig:visual}
\end{figure}

To generate DOGMs, we incorporate the Conditional Monte Carlo Dense Occupancy Tracker (CMCDOT) \cite{rummelhard2015conditional}. This approach  
associates four occupancy states to the grid. Each cell carries the probabilities of the cell being i) occupied and static, ii) occupied and dynamic, iii) unoccupied or free and iv) if the occupancy is unknown. The probabilities of these four states sum to one. 
In our work, we make use of three of these states and represent the grid as an RGB image. The channels Red, Green and Blue represent the unknown state, dynamic state and static state respectively. The associated probabilities of the cell in the 3-channel DOGM grid are interpreted as the pixel values of the RGB images. The RGB grid images can be seen in Fig. \ref{fig:overview}-\ref{fig:visual}.
Low probabilities in all three channels leave the grid-image black, therefore, representing free space. 

For allo-centric grid generation, we define the grid in the world frame, close to the initial position of ego-vehicle.
The state probabilities are initially computed in an ego-centric grid, since we use the on-board sensor data.   To ensure that we have cell information for the complete allo-centric grid dimensions when the vehicle is dynamic and moving away from the world frame origin, a much larger ego-centric DOGM is computed.
This information is then fused to update every cell states in the allo-centric grid in the world frame.

We compare the allo-centric and ego-centric grids at 4 time instants for the same scene and same grid dimensions in Figure \ref{fig:visual}.
In the allo-centric grid, the ego-vehicle (illustrated in the pink box) can be seen moving with respect to the grid, while it remains fixed in the ego-centric grid.
It is important to note that the observable space around the ego-vehicle remains the same for both grids. However, since they are defined in different frames, the two cover different spaces in the scene at a given time. 
We illustrate the common space covered by both grids since the start of the sequence, marked by yellow boundary.

\subsection{Problem Formulation}\label{sec:approach-b}

We formally define the task of predicting the scene in allo-centric DOGM representation, as sequence-to-sequence learning, see Fig. \ref{fig:overview}.
A sequence comprises of a set of sequential grid images that capture the evolution of a given scene. 
Let $X_t \in \mathrm{R}^{3 \textrm{x} W \textrm{x} H}$ 
and $Y_t \in \mathrm{R}^{3 \textrm{x} W \textrm{x} H}  $ be the $t$-th frame of the 3-channel grid-image where W and H denote the  width and height respectively. 
The input sequence for the grid-image is denoted by $X_{t-{N}:t}$, representing $N$ consecutive frames.
Given a set of input sequence, the task of the network is to predict future grid images, i.e. output sequence. 
The target and predicted output sequences are denoted by ${Y}_{t+1:t+{P}}$ and $\hat{Y}_{t+1:t+{P}}$ where $P$ is the prediction horizon.

For training and testing data, the DOGMs can be generated for both the input and the target sequences, leaving behind no additional need for labelled data or human intervention.
Since the input sequences, $X_{t-{N}:t}$, and output sequences, ${Y}_{t+1:t+{P}}$, are represented as images, 
this prediction task can be considered a video prediction problem.

\subsection{Deep Learning Prediction Architectures}\label{subsec:NN-architectures}
To study and compare the scene prediction with ego-centric and allo-centric grids, we train our datasets with different video prediction networks.
We consider 3 networks, briefly discussed in section  \ref{sec:related_work_a}: PredNet, PredRNN, LMC-Memory with memory alignment learning (here on referred as LMC-Memory). 

PredNet \cite{Lotter2017}, inspired from predictive coding, makes predictions based on how the predicted frames deviate from the target \cite{rane2020prednet}. 
The original work tests the network on vehicle mounted camera images from Kitti dataset \cite{Geiger2013IJRR} and demonstrates the ability to capture both egocentric motion as well as motion of objects in camera images.
We consider PredRNN \cite{9749915} and LMC-Memory architecture \cite{lee2021video} as the state of the art video prediction networks
that aim to capture long-term dependencies and motion context.
PredRNN implements novel ST-LSTM units with a zigzag internal memory flow and proposes memory decoupling loss to discourage learning redundant features. 
LMC-Memory architecture, on the other hand, proposes an external memory block with its own parameters to store various motion contexts.
The approach also offers an efficient computation method since the motion context for long-term multi-step predictions is computed only once for a given input sequence.  


We study these networks capabilities to retain the occupancy of the static region, and the ability to predict motion of dynamic agents in DOGM.


\subsection{Unknown Channel and Loss functions}\label{subsec:loss}\label{sec:approach-c}
In both ego-centric and allo-centric grids, a significant part of the scene remains unobserved, see Fig. \ref{fig:visual} (unknown channel is represented in red). This is more pronounced in the initial frames of the allo-centric grid, where the Lidar is unable to detect the farthest area from the ego-vehicle.

While it is more relevant to learn the evolution of static and dynamic components in the scene, inclusion of unknown channel is useful for our prediction task. A Lidar based grid is often unable to capture the full shape of a vehicle. For example, we can see in Fig. \ref{fig:visual} how the occupied cells by the bus vary in different time steps on the grid. It is only in the 2.0s time step that a rectangular shape is observed, otherwise different parts of the bus remain unknown.
The unknown channel at different instants also carries spatial information of the agents with respect to the ego-vehicle. Thus, with the sequential frames and the unknown channel, we assist the network to be able to extract spatial information and learn scene representation. 

The inclusion of unknown channel and emphasis on learning static and dynamic components is addressed in the loss function. 
Loss function $L$ in the implemented video prediction networks is modified to carry the weighted sum of the RGB channels:

\begin{equation}\label{eq:loss}
L = \alpha L_R +  \beta ( L_G + L_B )
\end{equation}
where,

$L_R$, $L_G$ and $L_B$ represent the loss for unknown (red), dynamic (green) and static channels (blue) respectively. In order to encourage the network to learn and improve the prediction of the static and dynamic channels, $\alpha$ is always kept smaller than $\beta$.


\section{Experiments} \label{sec:implementation}
 
\subsection{Dataset}\label{subsec:experiments}
We study the prediction performance on the real-world NuScenes dataset \cite{nuscenes2019}.
The original dataset consists of 850 scenes for training and 150 scenes for testing, 
each scene is approximately 20s long.
We generate the DOGM grid-based on the Lidar pointcloud and available odometry.
For allo-centric grid, we represent the scene with respect to a fixed reference frame and a grid dimension of 60 x 60m, with a resolution of 0.1m per cell. 
Each sequence starts with the ego-vehicle heading facing up, capturing the scene 10m behind and 50m ahead of it.   
The initial pose was selected to ensure that the ego-vehicle remains within the grid for the total sequence length, even when running at a high speed. 
For egocentric grid, we generate a grid of the same dimensions and resolution, and the ego-vehicle fixed in the center.
Each sequence is comprised of 35 frames, a time duration of 3.5s with DOGM grid images generated every 0.1s.
In total, we have 4,250 training and 750 testing sequences respectively.


\subsection{Training}\label{subsec:results}
The input sequence $X_{t-{9}:t}$ consists of 10 frames (1.0s). Each network is trained to make predictions $\hat{Y}_{t+1:t+{25}}$ for 25 future frames (2.5s).
Both the allo-centric and ego-centric datasets are trained with the original parameters of the respective video prediction network. 
For training with PredRNN and LMC Memory networks, both allo-centric and ego-centric grid images are resized to 192x192 pixels.
\textbf{PredRNN} is trained with a batch size of 4 and a learning rate of $10^{-4}$. The number of channels of each hidden state is set to 64. The loss function is the sum of L2 and decoupling loss, and the values of $\alpha$ and $\beta$ in Eq. (\ref{eq:loss}) are set to 0.2 and 0.8.
\textbf{LMC-Memory} is trained with a learning rate of $ 2$x$10^{-4}$, memory slot is set to 100 and ConvLSTM to 4 layers for frame predictions. The loss function is the sum of L1 and L2 losses. The values of $\alpha$ and $\beta$ are set to 0.2 and 0.8.
For training with \textbf{PredNet}, the grid images are resized to 160x160 pixels. The network is set to 4 hierarchical layers with an initial learning rate of $10^{-3}$.  The loss function is the L1 loss of only the first layer, the values of $\alpha$ and $\beta$ are set to 0.05 and 0.8.
All models are trained on Adam optimizer for 30 epochs.

\section{Evaluation} \label{sec:results}

\begin{figure*}[ht]
	\centering
	    \includegraphics[trim={5 5 5 5},clip,width=1.7\columnwidth]{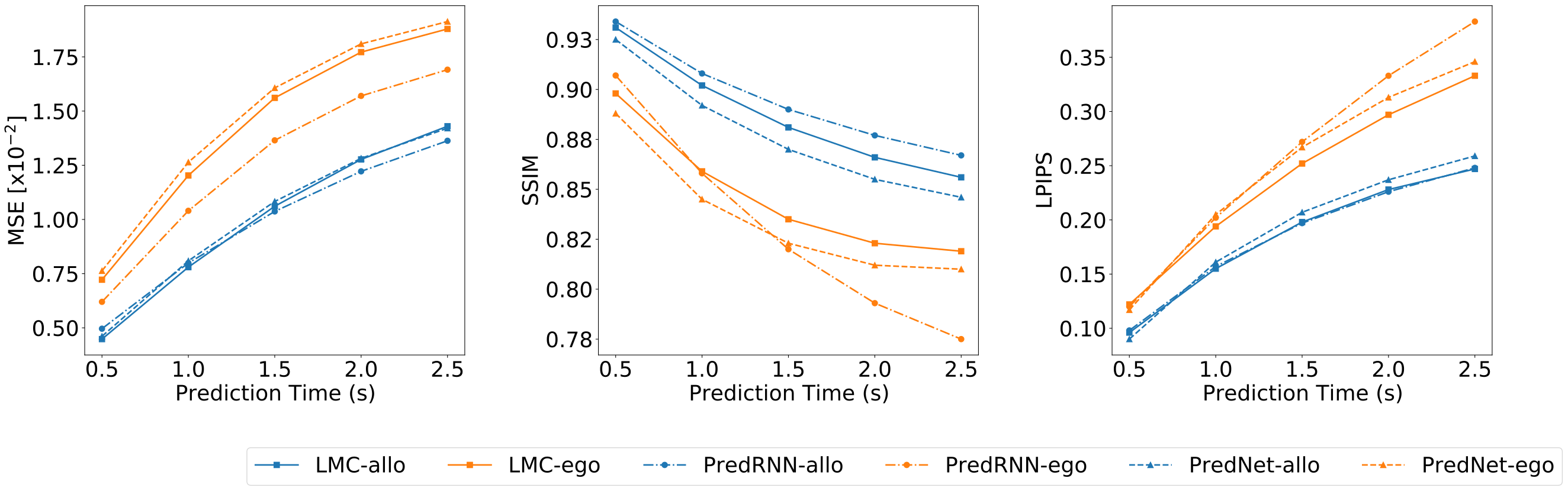}
\caption{Results with the MSE($\downarrow$), SSIM($\uparrow$) and LPIPS($\downarrow$) metrics with allo-centric and ego-centric grids for input sequences of 1.0s and prediction horizon up to 2.5s. For fair comparison, all test sequence frames were modified to only contain the scene observable in both the allo-centric and ego-centric grids. The allo-centric grid (results plotted in blue) outperforms the other with all three video prediction networks.}
\label{fig:quantity}
\end{figure*}

For evaluation, we are particularly interested in static and dynamic agents in the scene. We discussed in section \ref{subsec:loss}, the utility of unknown regions in learning scene representation.
But the unknown region occupies a big portion of the grid and, thus, in evaluation, overshadows the performance of more interesting and relevant segments: static and dynamic regions.
For this reason, we evaluate the dataset and network performances based on two channels of the predicted images, the blue and green channels representing static and dynamic components in the scene. 
We encourage the readers to refer to the video\footnote{\url{https://youtu.be/z-0BVM93X8c}} for a better visualization of the results.

\subsection{Quantitative Evaluation}\label{subsec:qntyevaluation}
The allo-centric and ego-centric grids at any instant observe different parts of the scene, see Fig. \ref{fig:visual}. For fair comparison between them, we modify the test dataset and crop out the part of each $t$-frame that has not been observed until the $t$-th sequence by both grids. Thus, for example, the part of the grids outside of the yellow dotted boxes in Fig. \ref{fig:visual} are blacked out for the input sequence frames $X_{t-{N}:t}$ as well as the target frames in the output sequence ${Y}_{t+1:t+{P}}$.

We measure the performances using three metrics:  MSE (Mean Square Error), SSIM (Structured Similarity Indexing Method), and LPIPS (Learned Perceptual Image Patch Similarity) \cite {zhang2018perceptual}. MSE is calculated by the pixel-wise difference between the ground truth and the predicted frame per channel and per cell. However, with MSE, the slightest error in predicted motion can result in large errors in the ego-centric grids dataset. The SSIM and LPIPS metrics evaluate the prediction results based on the structural similarity and perception similarity respectively. 
Lower values are better for MSE and LPIPS while  higher values are better for SSIM.

Table \ref{table:results} shows average results for the complete 2.5s prediction horizons. The MSE score of allo-centric grids is significantly lower compared to the one of ego-centric grids.
Since the complete scene transforms with respect to the ego-vehicle, the MSE is always higher in the ego-centric grid. The SSIM and LPIPS scores are also significantly superior for the allo-centric grid, due to the tendency of ego-centric grids to get increasingly blurry for higher prediction horizons.

\begin{table}[h]
\centering
\begin{tabular}{c|c|c|c|c|l|c|l|l|l|}
\cline{1-7}
\hline
\multicolumn{1}{c}{\textbf{Network}} & \multicolumn{2}{c}{\textbf{MSE} x $10 ^{-2}(\downarrow)$} & \multicolumn{2}{c}{\textbf{SSIM}$(\uparrow)$}                    & \multicolumn{2}{c}{\textbf{LPIPS}$(\downarrow)$}              \\\hline 
\cline{1-7} 
\multicolumn{7}{c}{\textbf{Allo-centric grid}}              \\ \cline{1-7} 
\cline{1-7} 
\multicolumn{1}{c}{LMC-Memory}                                 & \multicolumn{2}{c}{0.894} & \multicolumn{2}{c}{0.895} & \multicolumn{2}{c}{\textbf{0.167}}  \\
\multicolumn{1}{c}{PredRNN}                               & \multicolumn{2}{c}{\textbf{0.882}} & \multicolumn{2}{c}{\textbf{0.904}}  & \multicolumn{2}{c}{\textbf{0.167}}  \\ 
\multicolumn{1}{c}{PredNet}                               & \multicolumn{2}{c}{0.905} & \multicolumn{2}{c}{0.888}  & \multicolumn{2}{c}{0.172}  \\ \hline
\multicolumn{7}{c}{\textbf{Ego-centric grid}}              \\ \cline{1-7} 
\multicolumn{1}{c}{LMC-Memory}                               & \multicolumn{2}{c}{1.302} & \multicolumn{2}{c}{0.856}  & \multicolumn{2}{c}{0.217}  \\ 
\multicolumn{1}{c}{PredRNN}                               & \multicolumn{2}{c}{1.138} & \multicolumn{2}{c}{0.845}  & \multicolumn{2}{c}{0.234}  \\ 
\multicolumn{1}{c}{PredNet}                               & \multicolumn{2}{c}{1.335} & \multicolumn{2}{c}{0.847}  & \multicolumn{2}{c}{0.225}  \\ \hline
\end{tabular}
\caption{Average results with allo-centric and ego-centric grids for prediction horizon of 2.5s. The allocentric grid outperforms the other in all three video prediction networks.}
\label{table:results}
\end{table}

In Fig. \ref{fig:quantity}, we plot scores of the metrics for every 0.5s prediction step. The results with allo-centric grid (shown in blue) always perform better than the ego-centric grids. Among the three prediction networks, overall PredRNN performs the best with allo-centric grids. However, with the ego-centric grids (results shown in orange), PredRNN offers a good MSE score but the SSIM and LPIPS performances drop after 1.0s. This is because PredRNN tends to make blurry and diffused predictions in the output frames; this helps reduce the MSE but the scene loses its structures. This is further seen in the qualitative results discussed in section \ref{subsec:qltyevaluation} and illustrated in Fig. \ref{fig:quality}.

\subsection{Qualitative Evaluation}\label{subsec:qltyevaluation}

\begin{figure*}[h]
	\centering
	\hspace{10mm}
	\begin{subfigure}[b]{0.8\columnwidth}
		\centering
		\includegraphics[width=0.8\columnwidth]{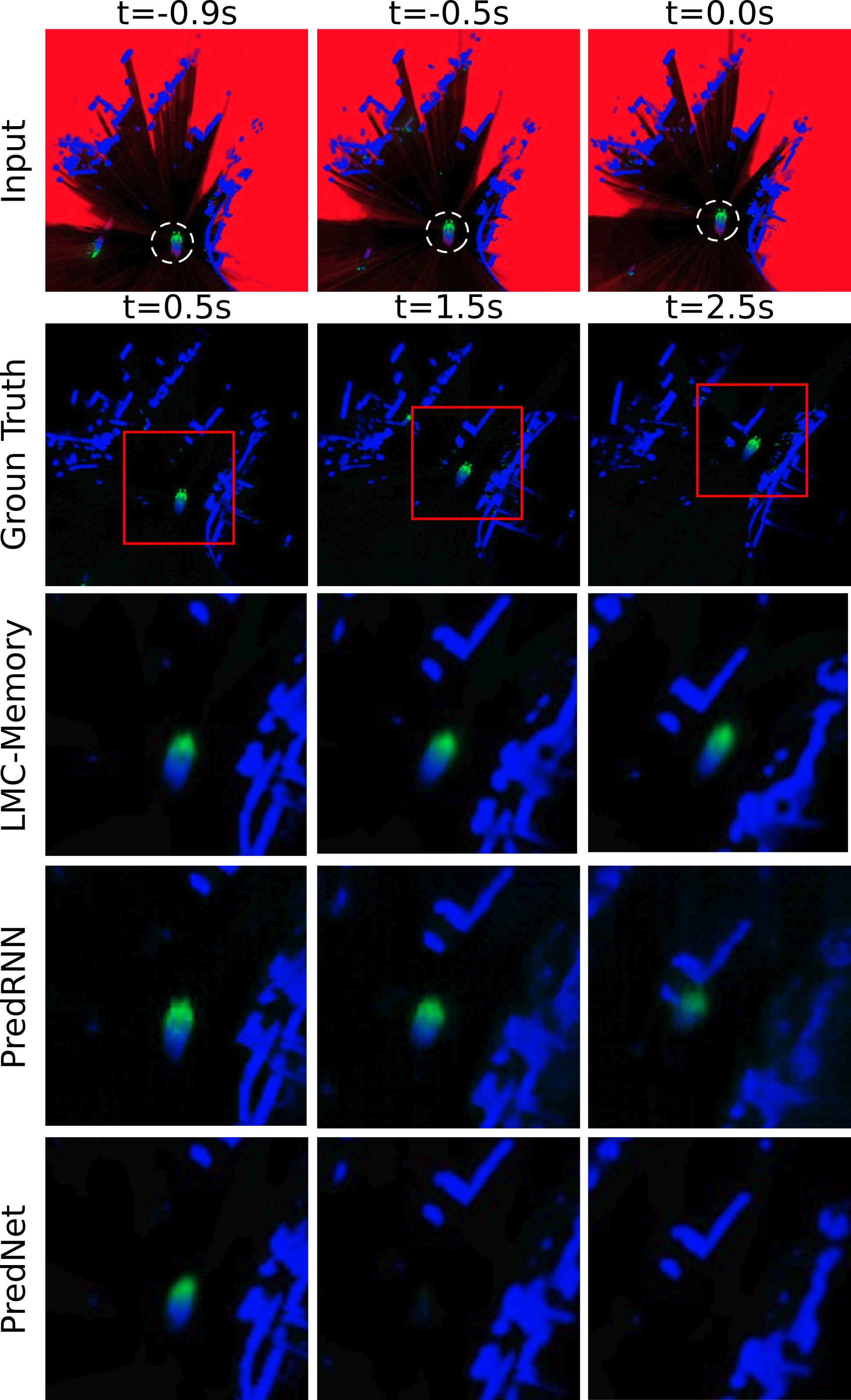}
		\caption{Allo-centric grids}
		\label{subfig:quality-allo}
	\end{subfigure}
	\hfill 
	\begin{subfigure}[b]{0.8\columnwidth}
		\centering
	    \includegraphics[width=0.8\columnwidth]{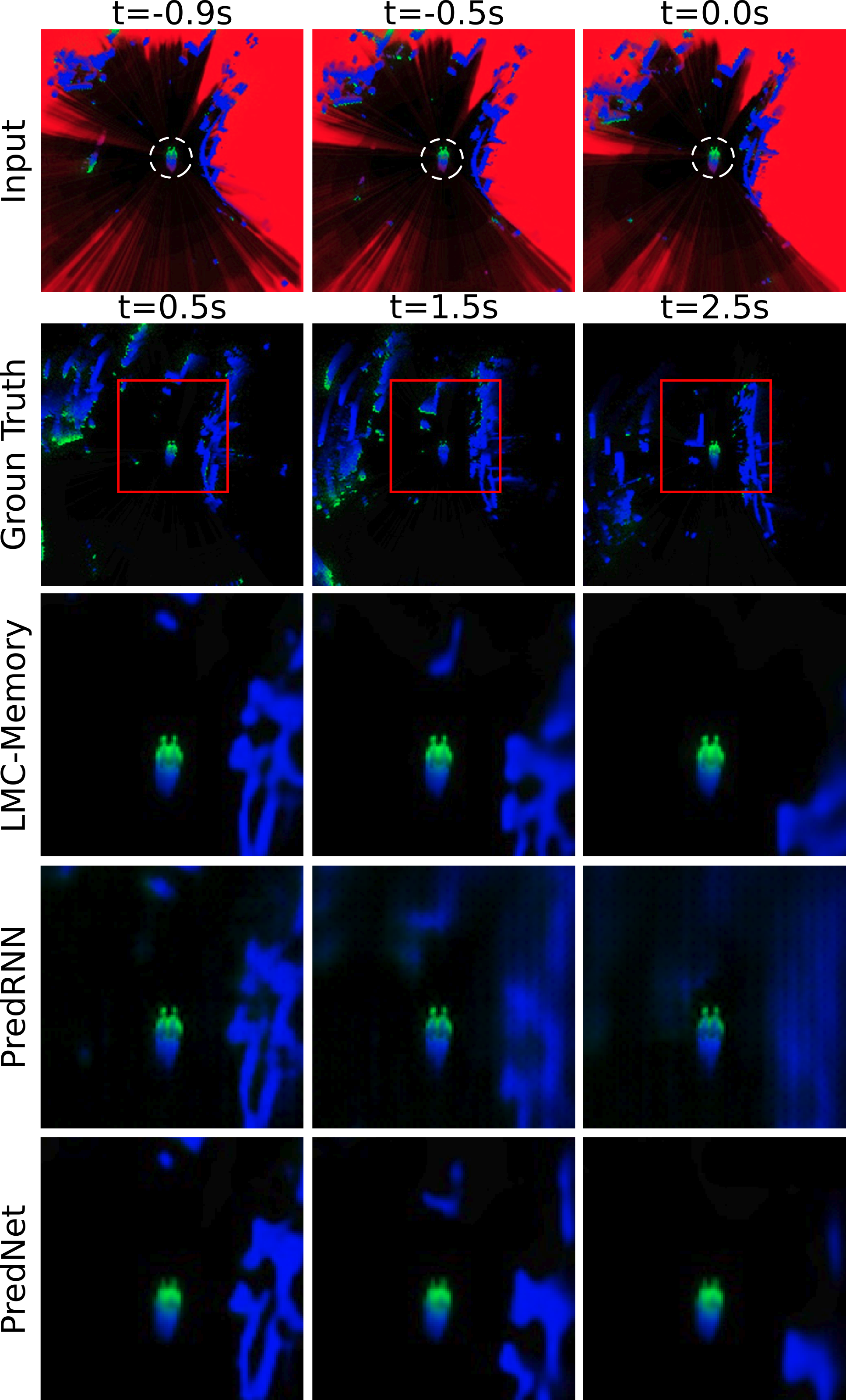}
		\caption{Ego-centric grids}
		\label{subfig:quality-ego}
    \end{subfigure}
	\hspace{10mm}
	\hfill 
\caption{Qualitative results for the ego-vehicle leaving a roundabout on both allo-centric (\ref{subfig:quality-allo}) and ego-centric  grids (\ref{subfig:quality-ego}). The input sequence consists of 10 frames (1.0s) and output predicted sequence of up to 25 frames (2.5s). The prediction results are shown at 0.5s, 1.5s and 2.5s instants and are magnified at the interesting spaces, marked by red box in the target(ground truth) frames. The best results can be observed with LMC-Memory network with the allo-centric grid that retains the scene structure and predicts the motion of the ego-vehicle best. }
\label{fig:quality}
\end{figure*}

The prediction results between the allo-centric and ego-centric grids differ drastically when the ego-vehicle is turning at an intersection or driving on a curved road. Figure \ref{fig:quality} shows results for a sequence where the ego-vehicle is exiting a roundabout. In this scene, while there are no other dynamic agents, the network needs to predict the behaviour of the ego-vehicle, when it is driving along the curved static segment (alluding to road structure) and is headed towards static objects/obstacles. 
For the allo-centric grids, the challenge is to predict the ego-vehicle pose while the scene remains static. The best results are achieved with the LMC-Memory. The vehicle pose is well-predicted up to 2.5s, its orientation is adjusted so that it does not hit the static components. For the same grid, the PredRNN fails to learn and predict the behaviour resulting in false prediction of collisions. The ego-vehicle, while getting more blurry, diffuses into the static obstacles on the road. With the PredNet, the ego vehicle is almost already lost at 1.5s prediction horizon. This is expected behaviour since PredNet is ideally not aimed at long-term video predictions. With all three networks, the ego-vehicle gets more blurry, however with PredRNN, the static scene also tends to get blurry at larger prediction horizon.

In the ego-centric grid, the whole scene rotates around the ego-vehicle. LMC-memory and PredNet significantly lose the static components ahead of the vehicle. The rotation results in increasing blurriness at every time step. PredRNN predictions are more diffused and faint blurry cells are still visible ahead of the vehicle, even at 2.5s prediction horizon. In context of planning and safe navigation, this high uncertainty in the environment structure renders the prediction results unreliable. 
\section{Discussion and Future Work}\label{sec:conclusion}
In this work, we presented a novel allo-centric dynamic ocuupancy grid approach for long-term prediction of urban traffic scene, and compared it to the conventional ego-centric DOGM approach.
We trained and tested various video prediction networks to show that allo-centric DOGM representation has superior ability to predict the same scene.

The most significant improvement is the allo-centric grid's ability to retain the static scene structure, especially when the vehicle turns. The ego-centric grid, on the other hand tends to lose the static scene, and hence the crucial information about whether the given space is occupied or free.

The results of allo-centric grids prediction with state-of-the-art PredRNN and LMC-Memory approaches have shown complementary benefits. PredRNN predictions, though diffuse and get more blurry, are capable of maintaining agents longer. 
We observe that LMC-memory shows better tendency at learning behaviours in comparison to the PredRNN.

It is pertinent to mention here that the two grids are still very similar. In both scenarios, the observable space updates relative to the position of the vehicle in the scene. Thus, in allo-centric grid while the grid is no more fixed to the ego-vehicle, the ego-vehicle bias remains.  

All three video prediction networks tested in this work address the prediction problem as deterministic. However, the behaviour of agents in urban traffic scene tends to be multimodal. For future work, the addition of multimodal prediction capabilities in the network architecture would be interesting. Additionally, the incorporation of semantics in the occupancy grid such as agent type and offline road information could assist in learning behaviours and interactions.

\bibliographystyle{IEEEtran}
\bibliography{main}%

\end{document}